\documentclass[runningheads,a4paper]{llncs}

\usepackage{amssymb}
\usepackage{graphicx}
\usepackage{subfigure}
\usepackage{multirow}
\usepackage{hhline}
\usepackage{longtable}
\usepackage{color}

{

\begin{document}

\mainmatter  % start of an individual contribution

\title{Predicting Crime Using Spatial Features}

\titlerunning{Predicting Crime Using Spatial Features}


\author{Fateha Khanam Bappee \inst{1} \and Am\'ilcar Soares J\'unior\inst{1} \and Stan Matwin\inst{1,2}}

%\author{}
\authorrunning{Bappee et. al. 2018}


\institute{Institute for Big Data Analytics, Dalhousie University, Halifax \and 
Institute for Computer Science, Polish Academy of Sciences, Warsaw}
%\institute{}


\toctitle{Predicting Crime Using Spatial Features}
\tocauthor{Authors' Instructions}
\maketitle

\begin{abstract}

%Understanding the high volume of crime data is very challenging, and often it is very complicated to unravel the pattern of crime as well as criminal behavior. 
Our study aims to build a machine learning model for crime prediction using geospatial features for different categories of crime. 
The reverse geocoding technique is applied to retrieve open street map (OSM) spatial data. 
%This operation propagates valuable information of the location that is used to profile spatial features for crime prediction. 
This study also proposes finding $hotpoints$ extracted from crime $hotspots$ area found by Hierarchical Density-Based Spatial Clustering of Applications with Noise (HDBSCAN). 
A spatial distance feature is then computed based on the position of different $hotpoints$ for various types of crime and this value is used as a feature for classifiers. 
We test the engineered features in crime data from Royal Canadian Mounted Police of Halifax, NS. 
%Four crime types such as alcohol-related crime, assault, property crime and motor vehicle crime are compared with original and engineered spatial features. 
We observed a significant performance improvement in crime prediction using the new generated spatial features.

\end{abstract}

%\vspace{-0.4in}
\section{Introduction}

%\note{[TODO][Fateha] Describe crime in general. Later, focus in specific crimes statistics.}

%Crime is one of the well-known social problems that affect the quality of life and slows down the economy of the country. 
In recent years, with the availability of high volume of crime data, scientists have been motivated to pursue research in the field of crime and criminal investigations. 
%For police and law enforcement agencies, it is very challenging to analyze the increasing volume of crime data. 
Understanding the factors related to different categories of crimes and their consequences is particularly essential. 
The study shown in \cite{Poverty} applies the procedure of statistical analysis on violent crime, poverty, and income inequality and outlines that homicide and assault has more connection and correlation with poverty or income inequality than other crimes. 
The research found that crime in the real-world highly correlates with time, place and population which make the researcher's task more complicated \cite{FAR}. 
Moreover, this geographical and demographic information contain many discriminatory decision pattern \cite{Discrimination,BigData}.
Leveraging data mining and machine learning techniques with crime research offer the analysts the possibility of better analysis and crime prediction, as well as mining association rules for crime pattern detection. 

Our study aims to build a machine learning model to predict the relationship between criminal activity and geographical regions. 
We choose Nova Scotia (NS) crime data as the target of our study. 
%Crime statistics for NS in 2012 show that the violent crime and the property crime rates per 100,000 residents are 1365.45 and 3932.45 respectively \cite{StCan1}. 
%Another study of Statistics Canada presents that the national rate of heavy alcohol consumption in Canada is 17.4\% where Nova Scotian's surpasses the average by 4.9\% \cite{StCan}. 
%According to the 2014 annual report of NS Trauma Program, approximately 25\% of motor vehicle accidents and 28\% of homicide or assault are attributable to alcohol \cite{NStrauma}. 
We focus on four different categories of crime: (i) alcohol-related; (ii) assault; (iii) property crime; and (iv) motor vehicle. 
In this work, we focus on the creation of two spatial features to predict crime: (i) gecoding; (ii) crime $hotspots$.
%Geocoding allows researchers to find various kinds of location information immediately by computing boundaries and distances while crime $hotspots$ may indicate areas where a type of crime is more likely to occur. %with their geographic boundaries and features. 
%In this work, we use the results of geocoding queries as spatial features provided for classifiers.
%Another spatial feature engineered in this work uses the creation crime hotspots using a density-based clustering algorithm. 
%In this work, we used the Hierarchical Density-Based Spatial Clustering of Applications with Noise (HDBSCAN) \cite{Campello2013-HDBScan} to create crime hotspots. 
%The experimental results indicate a very significant improvement of classification metrics with these two spatial engineered features on Nova Scotia crime data.

The contributions of this work include how geocoding can be used to create features using OSM data and crime $hotspots$ are  created using a density-based clustering algorithm. Moreover, $hotpoints$ are extracted from the $hotspots$. We show using a real-world scenario that these two new features increase the performance of different classifiers for predicting four different types of crime. 

%The contributions of this work are the following:
%\begin{itemize}
%\item We show how geocoding can be used to create features using OSM data.
%\item Crime $hotspots$ are created using a density-based clustering algorithm, then $hotpoints$ are extracted from the $hotspots$. After, we use the $hotpoints$ as features for classifiers.
%\item We show using a real-world scenario that these two new features increase the performance of different classifiers for predicting four different types of crime. 
%\end{itemize}

%\note{[TODO][Fateha] Write about why would be important to predict crime.}

%The rest of the paper is organized as follows. Section \ref{sec:relatedWork} reviews the related work. 
%Section \ref{sec:Feat} provides the details of spatial features engineered to improve prediction of crimes in NS.
%After, in Section \ref{sec:Experiments}, the data source, data preparation activities and experimental results are presented. 
%Finally, Section \ref{sec:conclusions} presents some concluding remarks with some future research directions.
%\vspace{-10pt}
\section{Related Work}
\label{sec:relatedWork}

%The relationship between crime and various features has been studied in many scientific and criminology research.  
%Nowadays, researchers are allowed to use spatial information from the real world using Geographical Information Systems (GIS). 
%Likewise, demographic information is easily accessible from different statistical sources. 
%The use of temporal dynamics between neighborhoods and crimes has also been broadly noted in criminology. 
%Therefore, 
The existing work on crime prediction can be categorized into three different groups based on the features such as temporal, spatial and demographic aspects.

%\subsubsection{Temporal aspect} 
%This part of work mainly says about temporal feature of crime prediction. 
%Works that considers the temporal aspect of crime prediction is detailed below.
Bromley and Nelson \cite{british} reveal temporal patterns of crime to predict alcohol-related crime in Worcester city. 
They also provide valuable insight into the spatial characteristics of the alcohol-related crime. The authors examine the patterns of crime and disorder at street level by identifying $hotspots$.
%Ratcliffe \cite{Ratcliffe_2004} focuses on temporal dynamics of crime pattern detection. 
Ratcliffe \cite{Ratcliffe_2004} proposes three types of temporal and spatial $hotspots$ for crime pattern detection. 
The author also shows how the spatial and temporal characteristics combine through his $hotspot$ matrix. 
However, the author did not apply any machine learning strategy to predict crime.  

In \cite{2578}, the authors analyze four categories of crime data which include liquor law violations, assaults and batteries, vandalism, and noise complaints. 
Different categories of crime show different temporal patterns. 
Brower and Carroll \cite{2578} clarify crime movement through the city of Madison using GIS mapping. 
The authors investigate the relationships among high-density alcohol outlets and different neighborhoods.
%Their studies show that serious crimes happen at the bar closing time between 2 am and 3 am and less severe crimes happen between 11 pm and midnight.  
%\subsubsection{Spatial aspect}
%Next, we list the works focused on spatial features of crime incidents. 
Chainey et al. \cite{KDE} identify crime $hotspots$ using Kernel Density Estimation (KDE) to predict spatial crime patterns. 
Similarly, in another study, Nakaya and Yano \cite{Viscrime} create crime $hotspots$ with the help of KDE. 
However, they combine temporal features with crime $hotspots$ analysis. 
%The paper \cite{Ratcliffe_2004} proposes three categories of spatial $hotspo$t such as dispersed, clustered, and $hotpoint$. 

Nath \cite{Nath} employed a semi-supervised clustering technique for detecting crime patterns.
%\subsubsection{Demographic aspect}
%This part of research focuses on crime pattern detection using more demographic information and criminal profiling. 
%In the study \cite{FAR}, the authors apply a fuzzy association rule mining technique to detect community crime pattern. 
%For mining association rules, the authors mainly consider demographic features.
%such as population density, mean people per household, people in the urban area, people under the poverty level and people in dense housing with some other features. 
In \cite{Rudin}, the authors propose a pattern detection algorithm named Series Finder to detect patterns of a crime automatically. 
%They consider a pattern as a series of crimes committed by the same offender or group of offenders. 
%Traunmueller et al. \cite{socinfo} analyze footfall count from telecommunication data and find a correlation between crime and metrics derived from population diversity. 
In \cite{BDataCrime}, the authors study crime rate inference problem using Point-Of-Interest data and taxi flow data. 
Point-Of-Interest data and taxi flow data are used to enhance the demographic information and the geographical proximity correlation respectively. 
%To build the inference model their study mainly focus on Linear Regression and Negative Binomial Regression model.

None of these features reported in the section were used for predicting crime categories alongside crime pattern detection. In our research, we mainly focus on the spatial aspect of crime prediction. 
We use geocoding technique and crime $hotspots$ to generate new features. 
\section{Engineering Spatial Features}
\label{sec:Feat}

%This section discusses the details of spatial features created for crime prediction. 
%Section \ref{sec:Geocoding feat} describes how the geocoding process is used to extract geographic information to create spatial features and 
%Section \ref{sec:SpatialFeat} outlines the techniques for crime $hotspots$ detection. 
%\vspace{-10pt}
%\subsection{Geocoding}
%\label{sec:Geocoding feat}
Geocoding is the process of spatial representation of a location by transforming descriptive information such as coordinates, postal address, and place name. 
%When the geographic coordinates are converted to get a location description, it is defined as reverse geocoding. 
The geocoding process relies on GIS and record linkage of address points, street network and boundaries of administrative unit or region. 
For this work, we used geocoding to extract the spatial information from the crime data. 
The geocoder library written in Python, was used for geocoding services with the Open Street Map (OSM) provider. 
%Figure \ref{fig:geocode} presents the framework of geocoding process. 
%Every crime in our dataset contains geographical coordinates (latitude and longitude) that are given to the Nominatim tool. Then, the tool queries the OSM dataset and outputs some information from geographic points.
%\vspace{-15pt}
%\begin{figure}
%\begin{center}
%\includegraphics[scale = 0.48]{geocode}
%\caption{Geocoding Framework} \label{fig:geocode}
%\end{center}
%\end{figure}8
%\vspace{-20pt}
The output of the Geocoder package can be 108 types of location including pubs, bus stops, or hospitals from NS.
According to OSM documentation, all of these types are grouped into 12 categories including amenity, shop, office etc. 
We used both types of location and category as features to predict crime.
%Figure \ref{fig:process}(a) and (b) show, respectively, alcohol-related crimes where the output of the geocoding process returned as categories highway and amenity, and as type bus stop and pub. 
%Figure \ref{fig:process} (a) and (b) show respectively, 36 alcohol-related crimes happened on and around bus stops in Halifax city and 44 alcohol-related crimes in pubs of downtown Halifax. 
%The paper uses Tableau 10.4 to map the crime data and OSM as a background map.

%\begin{figure}
%\begin{center}
%\includegraphics[scale = 0.60]{busstop_pubs_crime}
%\caption{Crimes on and around bus stops in Halifax city (red circles) (a) and pubs in downtown Halifax (dark blue circles) (b)} \label{fig:bstop}
%\end{center}
%\end{figure}

%\begin{figure}[!h] 
%\caption{Crimes on and around bus stops in Halifax city (red circles) (a) and pubs in downtown Halifax (dark blue circles) (b).}
%\centering	
%	\subfigure[Bus stops crimes.]{\includegraphics[width=.75\textwidth]{bus_stops.png}}
%   \hspace{1em}
%	\subfigure[Pubs crimes.]{\includegraphics[width=.75\textwidth]{pubs_crime}}
%\label{fig:process} 
%\end{figure}

%\vspace{-10pt}
%\subsection{Clustering for Hotspot creation}
%\label{sec:SpatialFeat}

%In general, hotspots define the areas with high-value intensity. %Too vague argument
The second type of feature used in this work was the creation of $hotspots$.
$Hotspot$ analysis can emphasize the patterns of data regarding time and location of a geographic area. 
%In the study of crime, one of the critical issues involves the analysis of where crimes occur in general. 
For a crime analyst, the creation of $hotspots$ became very popular to identify high concentrated crime area. 
In this work, $hotspots$ are created and transformed into a feature to predict different crime types. 
The idea is to cluster crime data into regions with a high rate of occurrence of the same crime type. 
%As we want to group crime data by density, a reasonable choice is to select the DB-Scan \cite{Ester1996-DBScan} algorithm. 
%However, since this algorithm has a complexity of O($n^2$), we decided to use the Hierarchical Density-Based Spatial Clustering of Applications with Noise (HDBSCAN). 
We decided to use HDBSCAN \cite{Campello2013HDBScan} because of its complexity (O($n\log_n$)) and because it can handle data with variable density and eliminates the $\epsilon$ ($eps$) parameter of DBSCAN which determines the distance threshold to cluster data. 
%Moreover, it has a complexity of O($n\log_n$).
%The clusters can identify a $hotspot$ areas where different types of crime occur. 
%Instead of considering the whole dense area as a feature,.
%We extract one geographical point named $hotpoint$ for every $hotspot$ found by the HDSCAN. 
%This $hotpoint$ is determined by extracting the $hotspot$ center, averaging the geographical positions inside the area.
%Finally, we extract the location of new crime reports and compute the distances to all $hotpoints$ found in the data and select the shortest distance to a $hotpoint$. 
%This shortest distance to a $hotpoint$ is then used as a feature for crime prediction.
%The framework for engineering spatial distance feature is shown in Figure \ref{fig:distance}. 
%Distances from all crime points to hot points are calculated for feature engineering. 
%Spatial distance feature is categorized based on how close a crime point is from a hot point. 
%Haversine distance metric is used for distance calculation. 
In this work, we used the Haversine distance in both HDBSCAN and shortest distance to a $hotpoint$.
The haversine formula determines the shortest distance between two points on earth located by their latitudes and longitudes.
%In this equation, $r$ is average earth radius (6371 km), $l_1$, $l_2$ define latitudes of two points and $\lambda_1$, $\lambda_2$ define longitudes of two points.
%%%%%%%%%%%%%%%%
%\begin{equation} \label{eq:haversine}
%	distance = 2*r*arcsin(\sqrt[]{d})
%\end{equation}
%\begin{equation} \label{eq:d}
%	d = \sin^2{(\frac{l_2-l_1}{2})} + \cos{(l_1)}*\cos{(l_2)}*\sin^2{(\frac{\lambda_2-\lambda_1}{2})}
%\end{equation}
%Figure \ref{fig:hotspot_all} visualizes four different crime maps with their hot points in Downtown Halifax. The upper left map illustrates the current assault crime (blue dots) and the hot point (orange dot enclosed by a circle) based on the former assault crime. Similarly, the motor vehicle-related crime (blue dots) and the hot point (green dot) are depicted in the upper right map. The lower left map shows all alcohol-related crimes in the downtown area with their hot point (red dot). Likewise, all property related crimes and their hot point (deep brown dot) are included in the lower right map. In Figure \ref{fig:hotspot_distance}, red arrows show the distances between hot point and crimes. The figure only presents the property crime and its hot point in downtown Halifax as an example.  

\begin{figure}[!h] 
\caption{An overview of the crime $hotspots$, $hotpoints$ and distance to $hotpoint$ feature.}
\centering	
	\subfigure[Crime data.]{\includegraphics[width=.4\textwidth]{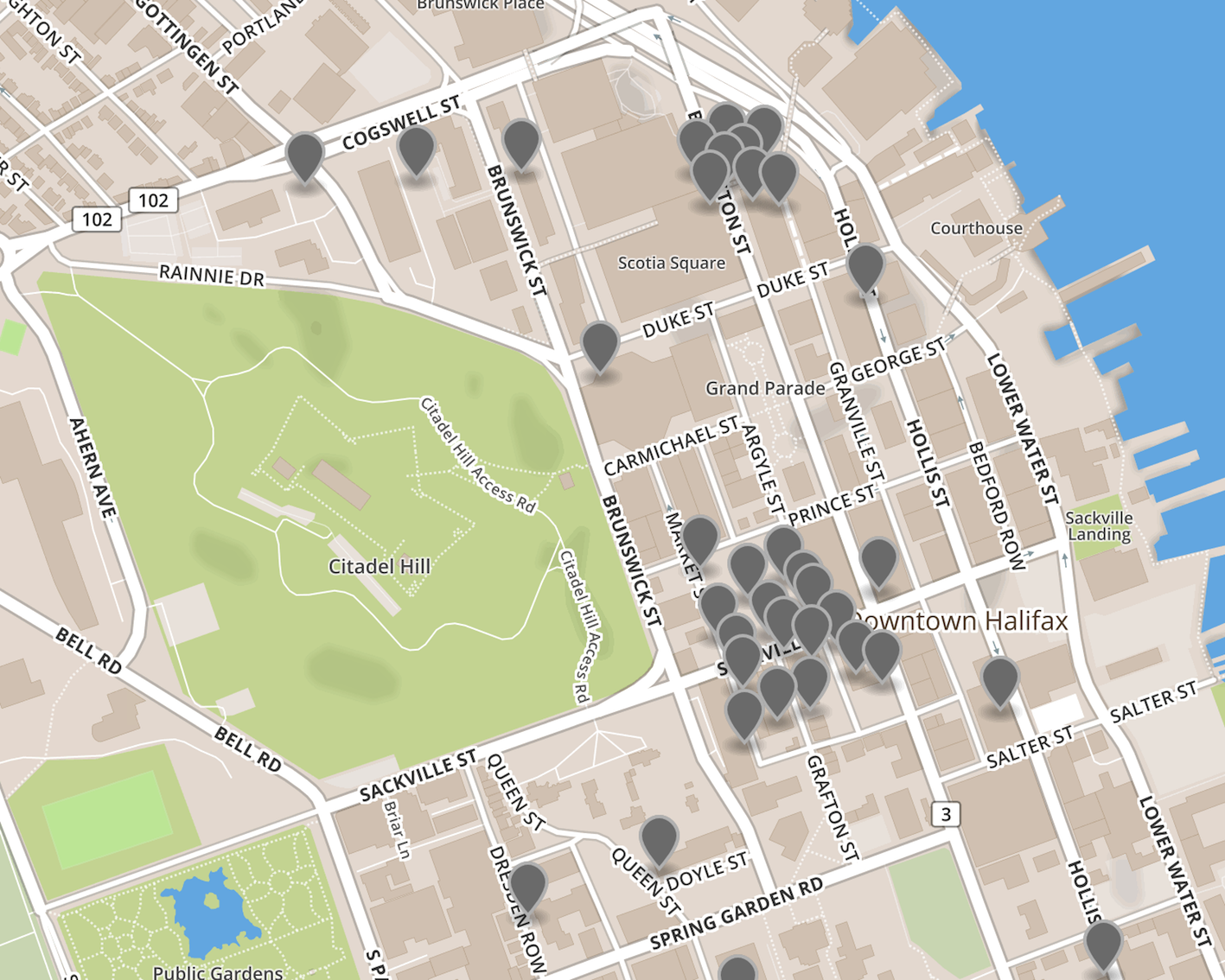}}
    \hspace{1em}
	\subfigure[A $hotspot$ created by HDBSCAN.]{\includegraphics[width=.4\textwidth]{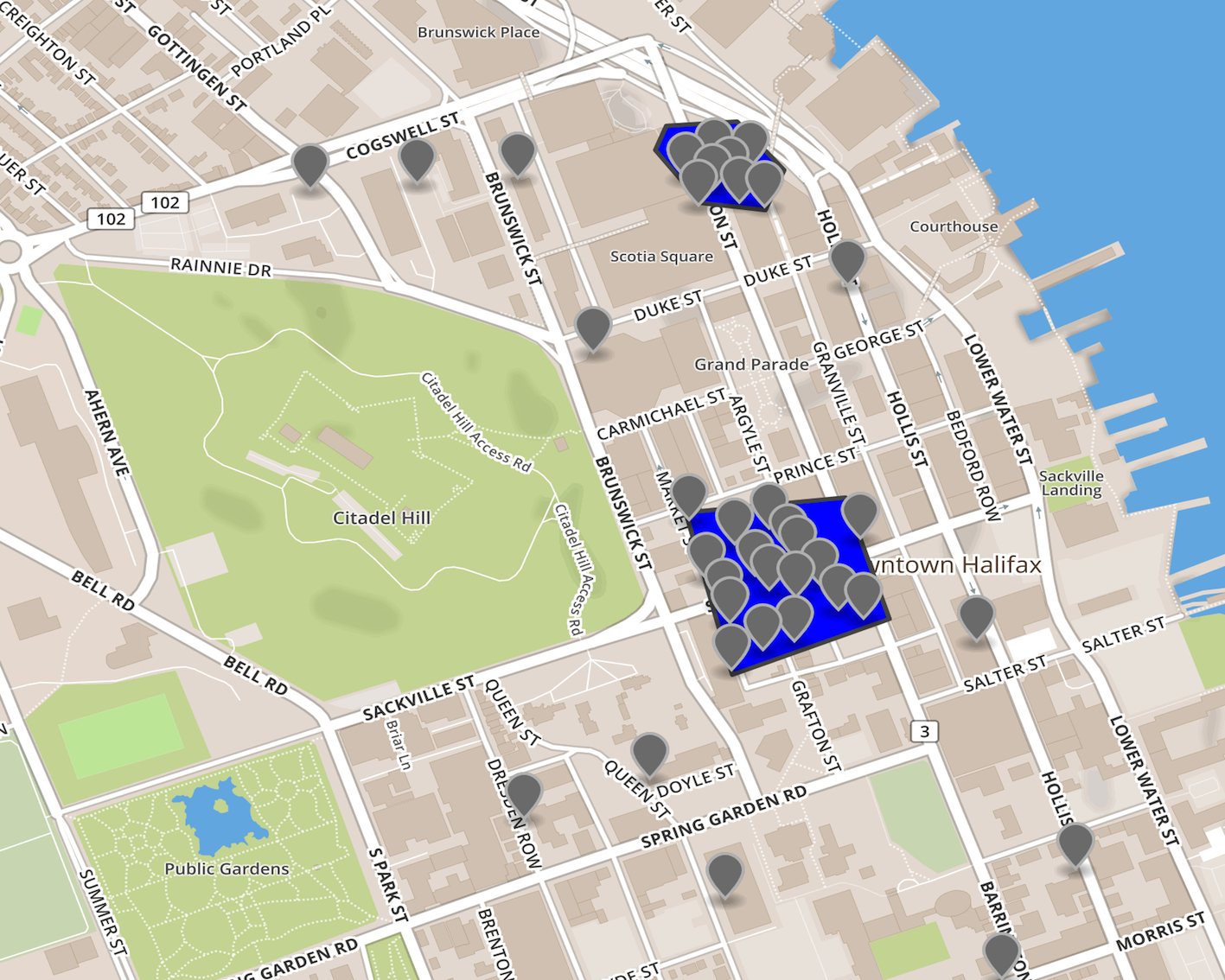}}
    \hspace{1em}
    \subfigure[A centroid computed from the $hotspot$.]{\includegraphics[width=.4\textwidth]{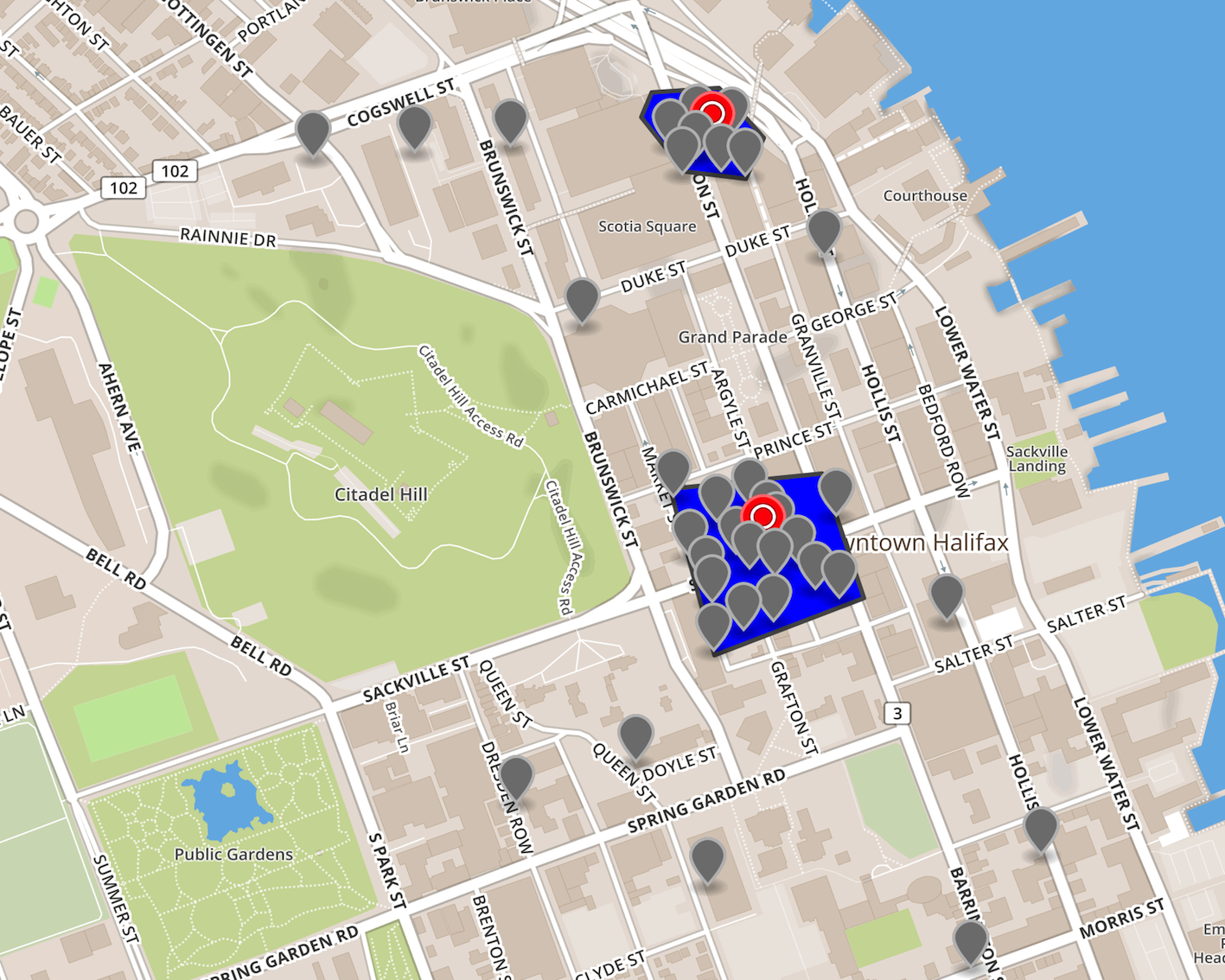}}
    \hspace{1em}
    \subfigure[Distance from centroid for new crime data around the $hotspot$.]{\includegraphics[width=.4\textwidth]{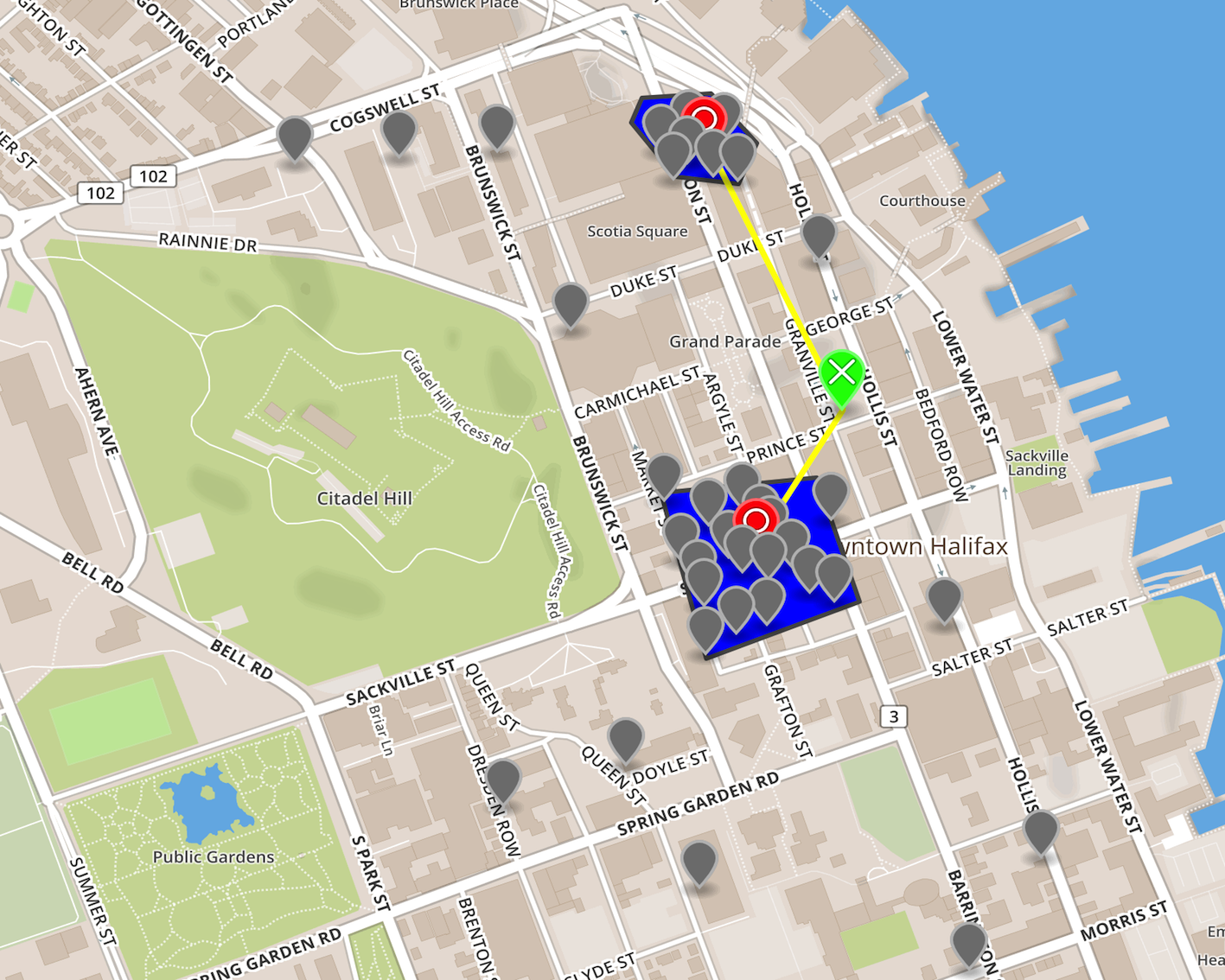}}
\label{fig:wholeprocess} 
\end{figure}

Figure \ref{fig:wholeprocess} summarizes the overall process to produce the shortest distance for $hotpoint$ feature.
Figure \ref{fig:wholeprocess} (a) shows crime examples (gray pins) in downtown Halifax area. 
Then, a $hotspot$ (blue area) found by HDBSCAN is shown in Figure \ref{fig:wholeprocess} (b).
Figure \ref{fig:wholeprocess} (c) shows a $hotpoint$ (red pin) extracted from a $hotspot$.
Finally, a new crime example (green pin) is evaluated, and the distances to $hotpoints$ (yellow line) are calculated. 
The feature used in this work will select the shortest distance to a $hotpoint$ as a feature for classifying a crime type.
%\vspace{-20pt}

%\vspace{-30pt}

%\begin{figure}
%\begin{center}
%\includegraphics[scale = 0.60]{distance_feature}
%\caption{Framework for spatial distance feature}  \label{fig:distance}
%\end{center}
%\end{figure}

%\begin{figure}
%\begin{center}
%\includegraphics[scale = 0.60]{hotspot_data}
%\caption{Hotspots in downtown Halifax based on four different types of crime}  \label{fig:hotspot_all}
%\end{center}
%\end{figure}

%\begin{figure}
%\begin{center}
%\includegraphics[scale = 0.50]{distance_hotspot}
%\caption{Distance from hotspot to property crime in downtown Halifax}  \label{fig:hotspot_distance}
%\end{center}
%\end{figure}

%\note{[TODO][Fateha] 1. What is a hotspot? 2. Why we used hotspots? 3. How we used hotspots? 4. What is the parallel between a hotspot and clustering? [important to have 1 or 2 figures explaining the technique.]}
\section{Experiments}
\label{sec:Experiments}

This section outlines the experiments performed in this work and reports the experimental results obtained by the proposed classifiers trained on all raw features and the engineered spatial features.
%Section \ref{sec:preparation} presents the source and retrieval of crime data, and how the data was prepared for the experiments. 
%Next, in Section \ref{sec:classifiers}, different learning algorithms used to predict crime are introduced. 
%Finally, Section \ref{sec:results} reports the experimental results obtained by the proposed classifiers trained on all raw features and the engineered spatial features described in this work. 

%\vspace{-10pt}
%\subsection{Data Source and Preparation}
%\label{sec:preparation}

Crime data from Halifax regional police department are used in this work, and it covers most of the districts in Nova Scotia province in Canada. 
%Our dataset was extracted from the Uniform Crime Reporting Survey (UCR).
%The UCR was designed to measure the incidence of crime and its characteristics in Canadian society. 
For our experiments, we explore all of the offenses of 2016 which include 3726 data samples. 
The crime attributes extracted from the source data include geographic location, incident\_start\_time, month, weekday, ucr\_descriptions, and whether the incident happened because of alcohol. 

We also group our data using four different classes, named alcohol-related, assault, property damage, and motor vehicle using the ucr\_descriptions and alcohol incident fields.
For the alcohol-related crimes, we considered all the cases where alcohol presence was reported in the UCR using the alcohol incident field (53\% alcohol, 47\% no alcohol).
For all the remaining classes, the ucr\_description field was used. 
%First, we consider the alcohol-related crimes class. 
%For this perspective, we have 1742 crimes that happened because of alcohol and 1984 crimes that have no relation with alcohol. %Info on the table
%Therefore, crimes those are related to alcohol are defined as positive class and otherwise negative class. 
%Next, we group all categories of ucr\_descriptions into three broad categories of crime, including assault, property damage, and motor vehicle. 
The assault group (65\% assault, 35\% no assault) covers all levels of assault including sexual assault, aggravated assault, bodily harm, threat, etc.
Property damage group covers break, theft, robbery, etc (65\% property damage, 35\% no property damage). 
Motor vehicle group covers all types of motor vehicle accident, act violation and impair driving (65\% motor vehicle, 35\% no motor vehicle). 
%The total data distributions with different categories of crimes and their ratio are summarized in Table \ref{tbl:datadesc}.
%For the alcohol-related group, 1742 crimes were reported as alcohol incident and 1984 crimes that have no relation to alcohol.
%The Assault group contains 1291 crimes related to assault and 2435 crimes with no relation to assault. 
%Next, in the Property Damage group, 431 crimes were reported as damage crimes and rest 3295 crimes were considered as not property damage crimes. 
%Finally, 686 crimes belong to motor vehicle group, and 3040 were grouped in the no motor vehicle-related group. 
%Positive and negative class of the crime data are labeled by an expert from Halifax Police Department. 

%\linespread{0.80}
%\begin{table}[ht]
%\centering
%\caption{Dataset description}
%\label{tbl:datadesc}
%\begin{tabular}{|l|c|c|c|}
%\hline
%\multicolumn{1}{|c|}{\textbf{Crime type}} & \textbf{Negative} & \textbf{Positive} & \textbf{Total} \\ \hline
%\textbf{Alcohol-related}                  & 1984(53\%)        & 1742(47\%)                 & 3726(100\%)    \\ \hline
%\textbf{Assault}                          & 2435(65\%)        & 1291(35\%)        & 3726(100\%)    \\ \hline
%\textbf{Property damage}                  & 3295(88\%)        & 431(12\%)         & 3726(100\%)    \\ \hline
%\textbf{Motor vehicle}                    & 3040(81\%)        & 686(19\%)         & 3726(100\%)    \\ \hline
%\end{tabular}
%\end{table}

To create the shortest distance to a $hotpoint$, we used UCR form data from the year of 2015. 
We created $hotspots$ for each positive class and the respective shortest distance to a $hotpoint$ was used in the experiment.
%For example, when the positive class was alcohol-related, we used a single shortest distance to a $hotpoint$ that was extracted from the examples of alcohol-related crime from 2015. 

%\linespread{0.75}
%\begin{table}
%\caption{Datasets used in the research}
%\begin{center}
%\begin{tabular}{l|l|l|l}
%\hline
%\textit{Data Source} & \textit{Total Documents} & \textit{Negative} & \textit{Positive}\\
%\hline
%\textit{Liquor Act} & 2649 & 1984 & 665\\
%\textit{Assault} & 3726 & 2435 & 1291\\
%\textit{Property Damage} & 3726 & 3295 & 431\\
%\textit{Motor Vehicle} & 3726 & 3040 & 686\\
%\hline
%\end{tabular}
%\end{center}\label{data}
%\end{table}

%\vspace{-10pt}
%\subsection{Evaluated Classifiers and Metric}
%\label{sec:classifiers}
The classifiers used in this work are Logistic Regression (LR), Support Vector Machine (SVM), and Random Forest (RF) and an Ensemble with all the previous classifiers. 
%Moreover, we considered Ensemble learning to obtain better predictive performance. 
%Ensemble learning combines multiple learning algorithms for this purpose. Model diversity is the key to creating a robust ensemble.
%An ensemble with the predictions of LR, SVM and RF methods are created to improve the performance of crime prediction. Equation \ref{ensem} gives the ensemble learning formulation of our problem.  
%\vspace{-5pt}
%\begin{equation} \label{ensem}
%  predictions \leftarrow (lr\_predictions + rf\_predictions*2 + svm\_predictions*2)/5  
%\end{equation}
%In this work, we used Scikit–Learn library \cite{scikit-learn} versions of LR, SVM, and RF to build models from crime data. 
We evaluate the classifiers' performance using the accuracy and Area Under the Curve (AUC) of the ROC (Receiving Operator Characteristic) analysis.
%We decided to use both metrics because the accuracy and the AUC complement each other.
The baseline used in this work to verify if the newly engineered features help a classifier to improve the crime prediction power was the raw data contained in the UCR form (incident\_start\_time, month, and weekday).
A 10-fold cross-validation was used in all phases to estimate model prediction performance correctly and paired t-tests (significance level of 0.05) were used to test the statistical difference significance of raw and engineered features. 
%The significance level of the paired t-test is 0.05. 
%According to the t-test, the engineered features show statistically significant performance i.e., the p-value is very small in all methods for Alcohol-related crime group.

%\vspace{-10pt}
%\subsection{Result Analysis}
%\label{sec:results}

%This section gives a detailed description of the experimental results for the impact of the features proposed in this work. 
Table \ref{tab:acc} shows the classification accuracy for LR, SVM, RF and an ensemble of these methods for all four categories of crime. 
For each method, the first column displays the accuracy of raw features and the second column for engineered spatial features. 
The * in Table \ref{tab:acc} symbol indicates that the method fails for the statistical hypothesis testing, i.e., the p-value is higher than 0.05.

%Needs more work in how to show the results.
For the Alcohol-related group, the results show that new spatial features achieve better accuracy in comparison with raw features for all four methods with statistical evidence support, and the Ensemble method performs better than others (75.52\% of accuracy) with almost 17\% accuracy improvement. 
%Paired t-tests are used to test the statistical significance between raw and engineered features. The significance level for the test is 0.05. 
%The t-tests states that the engineered features show statistically significant performance %(i.e., the p-value is smaller than 0.05) 
%in all methods for Alcohol-related crime group. 
The accuracy values of the engineered features for the Assault and Property damage groups shows that all methods, except LR, benefit from their inclusion. For example, adding engineered features with raw features improves nearly 11\% (Assault group) and 5\% (Property damage group) of accuracy for RF method.
Finally, for the Motor vehicle group, all the classifiers showed improvements, except for the Ensemble classifier.
%In Property damage and Motor vehicle groups, the accuracy of RF and SVM methods represent a significant performance by spatial features. 
%However, for Ensemble method, there is no statistically significant evidence of their performance. 
%Figure \ref{fig:acc} compares the accuracy to estimate the performance of methods with spatial features using bar graphs.   

\linespread{0.80}
\begin{table}[h]
\centering
\caption{Results for accuracy}
\label{tab:acc}
\begin{tabular}{l|c|c|c|c|c|c|c|c|}
\cline{2-9}
                                               & \multicolumn{2}{c|}{\textbf{LR}} & \multicolumn{2}{c|}{\textbf{RF}} & \multicolumn{2}{c|}{\textbf{SVM}} & \multicolumn{2}{c|}{\textbf{Ensemble}} \\ \hline
\multicolumn{1}{|c|}{\textbf{Crime type}}      & \textbf{raw}   & \textbf{eng.}   & \textbf{raw}   & \textbf{eng.}   & \textbf{raw}    & \textbf{eng.}   & \textbf{raw}      & \textbf{eng.}      \\ \hline
\multicolumn{1}{|l|}{\textbf{Alcohol-related}} & 59.36                & 65.27                & 57.73               & 73.51                & 59.28                & 71.31                & 58.61                  & 75.52                   \\ \hline
\multicolumn{1}{|l|}{\textbf{Assault}}         & 65.35               & 65.03*                & 47.94               & 58.89                & 63.53                & 65.27                & 55.96                  & 64.41                   \\ \hline
\multicolumn{1}{|l|}{\textbf{Property damage}} &                 88.43                & 88.41*               & 84.03                & 88.57                & 88.19                & 88.43               & 88.43   & 88.44*               \\ \hline
\multicolumn{1}{|l|}{\textbf{Motor vehicle}}   & 81.59               & 82.31                & 71.82               & 81.45                & 81.11                &  81.45               & 81.56                  & 81.80*                   \\ \hline
\end{tabular}
\end{table}

Table \ref{tab:auc} shows the AUC scores for LR, SVM, RF and an ensemble of LR, SVM \& RF methods. 
For Alcohol-related and Motor vehicle crimes, the results discovered that spatial features give better AUC scores than raw features for all four methods. For instance, the Ensemble method gives 82.5\%  and 69.4\% AUC score for Alcohol-related and Motor vehicle crimes respectively based on engineered features.
%Moreover, the performance is statistically significant according to the paired t-test. 
Similarly, for Assault and Property damage crime, LR, RF and Ensemble methods perform significantly better with engineered features. Adding engineered features with raw features gives 56.7\% and 65.7\% AUC score for Assault and Property damage crime respectively with the Ensemble method. Therefore, using spatial features, the Ensemble method performs at least 10\% improvement in AUC score for all four categories of crime. 
However, for SVM method, there is no significant evidence of improvement.

%\vspace{-30pt}
%\begin{center}
%\linespread{0.85}
%\begin{table}
%\caption{Accuracy based on LR, SVM, RF and an ensemble of LR, SVM $\&$ RF methods.}
%\begin{center}
%\begin{tabular}{l|l|l|l|l}
%\hline
%\textit{Crime Types (Accuracy) } & \textit{LR} & \textit{SVM} & \textit{RF} &\textit{Ensemble}\\
%\hline
%\textit{Liquor Act} & 71\% & 88\% & 88\% & 88\%\\
%\hline
%\textit{Assault} & 76.5\% & 77\% & 83\% & 88\%\\
%\hline
%\textit{Property Damage} & 77.5\% & 80.5\% & 86\% & 88\%\\
%\hline
%\textit{Motor Vehicle} & 77.5\% & 80.5\% & 86\% & 88\%\\
%\hline
%\end{tabular}
%\end{center}\label{accmodel1}
%\end{table}
%\end{center}
%\vspace{-28pt}
%%%%%%%%%%%%%%%%%%%%%%%%%%%%%%%
%\vspace{-30pt}
%\begin{center}
%\linespread{0.85}
%\begin{table}
%\caption{AUC based on LR, SVM, RF and an ensemble of LR, SVM $\&$ RF methods.}
%\begin{center}
%\begin{tabular}{l|l|l|l|l}
%\hline
%\textit{Crime Types (AUC) } & \textit{LR} & \textit{SVM} & \textit{RF} &\textit{Ensemble}\\
%\hline
%\textit{Liquor Act} & 71\% & 88\% & 88\% & 88\%\\
%\hline
%\textit{Assault} & 76.5\% & 77\% & 83\% & 88\%\\
%\hline
%\textit{Property Damage} & 77.5\% & 80.5\% & 86\% & 88\%\\
%\hline
%\textit{Motor Vehicle} & 77.5\% & 80.5\% & 86\% & 88\%\\
%\hline
%\end{tabular}
%\end{center}\label{aucmodel}
%\end{table}
%\end{center}

\begin{table}[h]
\centering
\caption{Results for AUC}
\label{tab:auc}
\begin{tabular}{l|c|c|c|c|c|c|c|c|}
\cline{2-9}
                                               & \multicolumn{2}{c|}{\textbf{LR}} & \multicolumn{2}{c|}{\textbf{RF}} & \multicolumn{2}{c|}{\textbf{SVM}} & \multicolumn{2}{c|}{\textbf{Ensemble}} \\ \hline
\multicolumn{1}{|c|}{\textbf{Crime type}}      & \textbf{raw}   & \textbf{eng.}   & \textbf{raw}   & \textbf{eng.}   & \textbf{raw}    & \textbf{eng.}   & \textbf{raw}      & \textbf{eng.}      \\ \hline
\multicolumn{1}{|l|}{\textbf{Alcohol-related}} &.575                & .723                 & .649                & .818                 & .635                & .747                & .661                   & .825                    \\ \hline
\multicolumn{1}{|l|}{\textbf{Assault}}         & .528               & .613                & .457               & .545                & .504                & .533*                & .459                  & .567                   \\ \hline
\multicolumn{1}{|l|}{\textbf{Property damage}} & .519               & .651                & .531               & .646                & .501                & .505*                & .534                  & .657                   \\ \hline
\multicolumn{1}{|l|}{\textbf{Motor vehicle}}   & .515               & .686                & .488               & .682                & .494                & .536                & .490                  & .694                   \\ \hline
\end{tabular}
\end{table}

%\begin{figure}
%\begin{center}
%\includegraphics[scale = 0.60]{chart_crime1}
%\caption{Accuracy} \label{fig:acc}
%\end{center}
%\end{figure}
\section{Conclusions and Future Work}
\label{sec:conclusions}

%Although many research and analysis of crime data are found in the literature, it is possible to verify that there is a lack of works that aim to create models for crime type prediction.
In this work, we explored the creation of spatial features derived from geolocated data. 
We created two types of spatial features:  (i) geocoding; and (ii) shortest distance to a $hotpoint$.
%The first used a geocoding service that can query OSM data and return a category and a type of information regarding where the crime occurred. 
%The second used the HDSCAN algorithm to create $hotspots$ grouped by type of crime, extracted a $hotpoint$ from each $hotspot$, and finally returned the shortest distance for a $hotpoint$ as a feature to feed a classifier. 
The new features were evaluated using four different crime types using only the information provided in the UCR forms as features for a classifier as the baseline. 
The results show that significant improvements in accuracy and AUC were found when the newly engineered features were added to the tested classifiers. 

We intend to extend this work in other directions.
%(i) integrate social and economic data to the current dataset and explore the performance of models when such information is available; and (ii) we want to explore some new and existing way (i.e. KDE) to create $hotspots$ and $hotpoints$. 
%The extracted knowledge from this exploration will be compared with what we have achieved using HDBSCAN.}  
As our study focuses on real world datasets, the subject of data discrimination is another important concern. Data discrimination refers to bias that happens because of contradistinction among different data sources. 
%Next, we will measure and address if there is any unexposed discriminatory decision pattern exists in the demographic and geographic crime data.
Another research direction we want to explore is the possibility of performing transfer learning from what was learned in NS to other Canadian provinces. 

\subsubsection*{Acknowledgments}
The authors would like to thank NSERC, NS Health Authority and Injury Free Nova Scotia for financial and other supports.

\bibliographystyle{splncs03}
\bibliography{Alcohol_Crime}

\end{document}